\documentclass[twocolumn, switch]{article}
\usepackage{preprint}
\usepackage{hyperref}
\usepackage[numbers,square]{natbib}

\usepackage{amsmath}
\usepackage{booktabs}
\usepackage{graphicx}
\hypersetup{colorlinks=true, linkcolor=purple, urlcolor=blue, citecolor=cyan, anchorcolor=black}

\usepackage[utf8]{inputenc}	% allow utf-8 input
\usepackage[T1]{fontenc}
\usepackage{xcolor}
\usepackage{lineno}					% Line numbers
\usepackage{tikz} 					% ORCiD insertion

\usepackage{newfloat}
\DeclareFloatingEnvironment[name={Supplementary Figure}]{suppfigure}
\usepackage{sidecap}
\sidecaptionvpos{figure}{c}

\usepackage{titlesec}
\titlespacing\section{0pt}{12pt plus 3pt minus 3pt}{1pt plus 1pt minus 1pt}
\titlespacing\subsection{0pt}{10pt plus 3pt minus 3pt}{1pt plus 1pt minus 1pt}
\titlespacing\subsubsection{0pt}{8pt plus 3pt minus 3pt}{1pt plus 1pt minus 1pt}

\definecolor{lime}{HTML}{A6CE39}
\DeclareRobustCommand{\orcidicon}{
	\begin{tikzpicture}
	\draw[lime, fill=lime] (0,0)
	circle [radius=0.16]
	node[white] {{\fontfamily{qag}\selectfont \tiny ID}};
	\draw[white, fill=white] (-0.0625,0.095)
	circle [radius=0.007];
	\end{tikzpicture}
	\hspace{-2mm}
}

\title{Self-Supervised Learning of Gait-Based Biomarkers}

\usepackage{authblk}

\author[1, 2\thanks{\texttt{rcotton@sralab.org}}]{R. James Cotton\href{https://orcid.org/0000-0001-5714-1400}{\orcidicon}}
\author[1, 3]{J.D. Peiffer\href{https://orcid.org/0000-0003-2382-8065}{\orcidicon}}
\author[1]{Kunal Shah}
\author[1]{Allison DeLillo}
\author[1]{Anthony Cimorelli}
\author[1]{Shawana Anarwala}
\author[1]{Kayan Abdou}
\author[1, 2]{Tasos Karakostas}
\affil[1]{Shirley Ryan AbilityLab}
\affil[2]{Northwestern University, Department of Physical Medicine and Rehabilitation}
\affil[3]{Northwestern University, Department of Biomedical Engineering}

\begin{document}

\twocolumn[\begin{@twocolumnfalse}

\maketitle

\begin{abstract}
Markerless motion capture (MMC) is revolutionizing gait analysis in clinical settings by making it more accessible, raising the question of how to extract the most clinically meaningful information from gait data. In multiple fields ranging from image processing to natural language processing, self-supervised learning (SSL) from large amounts of unannotated data produces very effective representations for downstream tasks. However, there has only been limited use of SSL to learn effective representations of gait and movement, and it has not been applied to gait analysis with MMC. One SSL objective that has not been applied to gait is contrastive learning, which finds representations that place similar samples closer together in the learned space. If the learned similarity metric captures clinically meaningful differences, this could produce a useful representation for many downstream clinical tasks. Contrastive learning can also be combined with causal masking to predict future timesteps, which is an appealing SSL objective given the dynamical nature of gait. We applied these techniques to gait analyses performed with MMC in a rehabilitation hospital from a diverse clinical population. We find that contrastive learning on unannotated gait data learns a representation that captures clinically meaningful information. We probe this learned representation using the framework of biomarkers and show it holds promise as both a diagnostic and response biomarker, by showing it can accurately classify diagnosis from gait and is responsive to inpatient therapy, respectively. We ultimately hope these learned representations will enable predictive and prognostic gait-based biomarkers that can facilitate precision rehabilitation through greater use of MMC to quantify movement in rehabilitation.\\
\end{abstract}


\vspace{0.5cm}

\end{@twocolumnfalse}]

%%%%%%%%%%%%%%%  Main text   %%%%%%%%%%%%%%%

\section{Introduction}\label{Introduction}

The development of markerless motion capture (MMC) has greatly lowered the barrier to the biomechanical analysis of movement. With this technology, it is now feasible to perform routine gait analysis on individuals undergoing rehabilitation for gait impairments, and even longitudinally during treatment. This presents a new set of challenges, as each gait analysis produces a huge amount of data including spatiotemporal gait parameters and whole-body joint kinematics down to individual fingers. To effectively utilize this technology in clinical care, it is essential to reduce each gait analysis into a much smaller set of summary statistics that can be reviewed by treating clinicians.  Ideally, these statistics should be interpretable, valid, reliable and robust, sensitive to clinically meaningful change, and clinically actionable. Biomarkers provide a useful framework to consider such statistics. For example, some types of biomarkers identified by an NIH workgroup are diagnostic biomarkers, which signal the presence of a disease, and response biomarkers, which measure the response to interventions \cite{fda-nih_biomarker_working_group_fda-nih_2021}.

This raises the question of how to identify these gait-based biomarkers. One approach for learning useful representations is self-supervised learning (SSL). SSL establishes a pretext task that does not require data manual labeling, where training a system to solve this task produces a representation is effective for other downstream tasks. This has been successful for a range of applications, including natural language processing, image categorization, scene segmentation, and mapping between images and language. There are many potential pretext tasks used in SSL. For example, predicting the next word is commonly used for SSL in language with great success. This can naturally be extended to gait by predicting a future trajectory from a set of observations.

\citet{winner_discovering_2022}  used a recurrent neural network to predict the next time step from marker-based gait analysis, which produced individualized gait signatures clustered by whether participants had a history of stroke or not and separated by degree of impairment. They also showed this network could autoregressively generate sample trajectories, which made the gait signatures more interpretable as a change in gait signatures could be visualized through a change in sampled kinematics. \citet{endo_gaitforemer_2022} used a transformer with an encoder-decoder architecture trained to predict a block of future timesteps movements estimated using human pose estimation from monocular videos. They pretrained the network on a large existing dataset from able-bodied individuals containing multiple actions, before fine-tuning it on a small dataset of videos from individuals with Parkinson's disease. In addition to the prediction pretext task, they trained a classification head on the input trajectory to predict the activity being performed in the pretext task and then fine-tuned this to predict clinical gait impairment scores.

Several other pretext tasks have shown success on a range of tasks that, to our knowledge, have not been applied to clinical gait analysis. Contrastive learning involves taking different augmentations of samples that should not change the underlying meaning (positive samples) and learning a representation that makes those augmented views close together in an embedding space while being far away from other samples \citep{oord_representation_2018}. The learned representation should embed more similar samples more closely together, despite never receiving explicit labels, which can produce a representation useful for downstream tasks. This has been applied with great success to learn visual representations that perform well on object classification \citep{chen_simple_2020}. The augmentations to produce positive samples should be designed to not change the relevant information in the sample. In the case of gait, temporal crops that take different observations of the same walking condition provide a natural approach. Another pretext task is similar to future prediction but involves masking some of the inputs and forcing the system to predict those missing samples \citep{devlin_bert_2019}. This commonly uses bidirectional transformers, which are challenging to integrate with causal prediction. However, Forgetful Causal Masking (FCM) can combine masking input tokens with a causal mask that predicts future tokens and substantially improves language model performance for downstream tasks \citep{liu_towards_2023}.

MMC can enable much greater use of gait analysis in clinical care and research, but using SSL to learn representations from this data has not been explored. The works described above used marker-based motion capture and human pose estimation applied to monocular video \citep{winner_discovering_2022, endo_gaitforemer_2022}. Several pretext tasks have not been explored for gait representation learning. This work aimed to use SSL to learn a good representation from gait data acquired with MMC in a rehabilitation hospital from a dataset that contains a mixture of diagnoses with a high representation of stroke survivors and lower limb prosthesis users. We also wanted to explore the influence of different neural network architectures and SSL objectives, including using contrastive learning. The utility of a learned representation is its performance on downstream tasks. We tested the performance of the learned representations on three downstream tasks: diagnosis, laterality of impairments, and monitoring changes in gait impairments during rehabilitation.

In short, the contributions of this work include:

\begin{itemize}
\item We show that contrastive learning, using different time windows from a given gait trial as positive examples, is a powerful objective for learning gait representations. Without any explicit annotation, it learns embeddings that are similar for different trials from a given participant.
\item By fine-tuning a simple linear decoder on these embeddings we can classify diagnoses (diagnostic biomarker). The embeddings of stroke survivors undergoing inpatient therapy also shift closer to control participants after therapy (response biomarker). These show how much health information can be inferred with gait analysis powered by SSL.
\item We show that SSL can be successfully applied to gait analysis data collected with MMC.
\item We explore a range of hyperparameters including network architecture, augmentations, input masking, pretraining on other datasets, and loss functions on the contrastive loss and downstream task performance.
\end{itemize}

\section{Methods}\label{Methods}

\subsection{Data collection}\label{Data collection}

The participants, acquisition system, and methods to perform inverse kinematic fits to estimate joint angles are described in the appendix. In short, gait analysis was performed with markerless motion capture on 75 individuals seen during inpatient and outpatient therapy as well as healthy controls. Of these, 18 were controls, 22 had a history of stroke, and 26 were lower limb prosthesis users, with the remainder having a variety of conditions. Walking over 7 meters was collected in a variety of conditions, such as different speeds and assistive devices. The dataset included 712 walking trials. Keypoints detected from multiple views were reconstructed with an implicit representation mapping from time to 87 3D marker locations \citep{cotton_improved_2023} and then inverse kinematic fits with a biomechanical model were performed on these markers to extract joint angles \citep{nimblephysics, werling_rapid_2022, cotton_markerless_2023}. For this study, we used the sagittal plane joint angles from the hip, knee, and ankles and also included back and elbow flexion.

\subsection{Network architecture, losses, and training}\label{Network architecture, losses, and training}

Gait trajectories were processed with a transformer \citep{vaswani_attention_2017},  specifically a decoder-only architecture based on GPT-2 \citep{radford_language_2019}, which consists of multiple self-attention and feed-forward neural network layers. Each layer in the network consists of a multi-head self-attention mechanism followed by a position-wise feed-forward neural network. The feed-forward network used a 4-time expansion of the hidden dimension. The self-attention mechanism used 4 attention heads. We performed tests with both learned positional embeddings and rotary encoding \citep{su_roformer_2022}. This was implemented in Jax using Equinox \citep{kidger_equinox_2021}. We used Haliax \citep{haliax} with named tensor dimensions and an architecture like GPT2 \citep{radford_language_2019},  based on the implementation from Levanter \citep{levanter}.

The input sequence length was 90. The output sequence tokens were mapped to future predictions, through a final linear layer. We used causal masking, which was sometimes combined with FCM. To decode the overall gait representation, we either extracted the last temporal token or used an additional token appended to the end of the sequence, called a [CLS] token. The input joint trajectory space was projected into the token embedding dimension with a linear layer. They were optionally augmented by adding noise proportional to the standard deviation of each joint over the entire training dataset (typically 10\% of the standard deviation). The network was trained with the AdamW optimizer from Optax for 1000 epochs with a learning rate of 2e-3 \citep{loshchilov_decoupled_2019, deepmind2020jax}. The network batch size was 32.

%  We used a cosine decay schedule with a learning rate decay every 10000 training steps with a final learning rate of 2e-5.

\subsubsection{Losses}\label{Losses}

For the contrastive loss, we projected the gait representation (using either the last token or an additional [CLS] token) into a 16-dimensional space with a linear layer. Like prior work, this was normalized to a length of one before and after projection \citep{chen_simple_2020}. Positive samples were two 90-frame (3 second) trajectories sampled from the same walking trial, which could potentially overlap. Negative samples were drawn from any other trial, including potentially another from the same session. We did not treat multiple trials from the same person and condition as positive examples, as we were trying to avoid using labels. The contrastive loss was computed as in Eq (\ref{eq:contrastive_loss}), where $z_i$ and $z_j$ are the embeddings of the positive samples and $z_k$ are the embeddings of the negative samples. $\tau$ is the temperature, which was fixed at 0.1. The similarity function was the cosine similarity. $N$ is the batch size before generating the two temporal crops from each trial.

\begin{equation}
\label{eq:contrastive_loss}
\mathcal{L}_{contrastive} = -\log \frac{\exp(\mathrm{sim}(z_i, z_j) / \tau)}{\sum_{k=1}^{2N} \mathbb{1}_{[k \neq i]} \exp(\mathrm{sim}(z_i, z_k) / \tau)}
\end{equation}

For the next timestep prediction loss, embeddings from the last layer were projected via a linear layer into the original joint space to produce predictions of the next time step, $t$, for each joint, $j$, $\hat{y}_{t,j} \in \mathbb R$. We used a mean-squared error loss on these predictions against the ground truth one sample into the future, $y_{t,j}$, Eq (\ref{eq:prediction_loss}). Note because of the future prediction, $y_{t,j} = x_{t -1,j}$ where $x_{t,j}$ is the input trajectory. Because $y_{t,j}$ is not defined for $i=0$ this time step was not included in the loss. Throughout this work, we used a constant sequence length of $T=90$ (3 seconds), and $J$ was the number of joints.

\begin{equation}
\label{eq:prediction_loss}
\mathcal{L}_{prediction} = \frac{1}{(T-1)J} \sum_{t=2}^{T}  \sum_{j=1}^{J} \left( \hat{y}_{t,j} - y_{t,j} \right)^2
\end{equation}

The contrastive loss was always included during training and the prediction loss was only included in some of the runs with a weighting producing a total loss in Eq (\ref{eq:total_loss}) using a $\lambda$ of either 1.0 or 0.0.

\begin{equation}
\label{eq:total_loss}
\mathcal{L} = \mathcal{L}_{contrastive} + \lambda \mathcal{L}_{prediction}
\end{equation}

We pretrained our network on the gait laboratory dataset for 1000 epochs using these losses before fine tuning for the tasks described below. When fine tuning, we discarded the contrastive projection head and joint angle prediction head.

\subsubsection{Hyperparameter testing}\label{Hyperparameter testing}

The dataset was split for 3-fold cross-validation based on subject identities (i.e., subjects in the validation test were never seen during training). For each hyperparameter (HP) setting, we trained the model and then the downstream classifiers for diagnosis and laterality. We then computed the prediction loss, contrastive loss, and performance on downstream tasks on the validation data. For each HP setting, we recorded the average cross-validated performance.

HPs adjusted included the number of layers, the transformer embedding dimensionality, the use of an additional [CLS] token for the contrastive loss, the use of a prediction loss, the use of noise augmentation, the use of forgetful causal masking,\texttt{1}  fixed sinusoidal positional encoding versus rotary encoding, and the number of epochs of pretraining. Experiments were tracked with Weights and Biases \citep{wandb}.

We initially experimented with automated HP optimization, but as discussed later, did not find that improving the contrastive loss translated into downstream task performance. Thus we opted for more of a semi-systematic exploration of different HP parameters.

\subsection{Downstream tasks}\label{Downstream tasks}

\subsubsection{Diagnosis classification}\label{Diagnosis classification}

To quantify the clinically relevant information contained in the SSL representation, we trained a linear readout for a logistic classifier to predict diagnosis and laterality. This received the input from the layer before the contrastive projection, similar to other works \citep{chen_simple_2020}. Because the hidden dimensionality was substantially larger than the number of samples, we also included L1 regularization. We performed inner 4-fold cross-validation over subjects in the training set while sweeping logarithmically over an L1 regularization weight from -6 to -3 to find the optimal regularization parameter. We then trained the classifier with this optimized regularization value on the learned embeddings from the whole training set before finally testing on the corresponding outer 3-folded validation set.

To be explicit, in each case, both the gait representation \textit{and} classifier were trained on only the training data, before first embedding the validation trajectories and scoring them. No subjects in the validation data were included in the training data (i.e., this was not cross-validation over trials). The classification was performed on only 3 seconds of data from a trial and for the validation trials, this was selected from the middle of a walking segment.

We used this approach to fit logistic regression classifiers for stroke versus control, lower-limb prosthesis user versus control, and impairment laterality. Given our sample size, this typically resulted in 6 to 9 people with each diagnosis in each validation set. We report absolute accuracy and did not adjust for any imbalances between training and validation sets.

\subsubsection{Biomarker for progress}\label{Biomarker for progress}

To explore whether this learned representation produces a meaningful topology, we tested its ability to quantify and detect changes during inpatient rehabilitation. We performed this analysis on a subset of our participants (n=11) with a recent history of stroke who were undergoing inpatient rehabilitation and had an MMC gait analysis both early and later during treatment. Due to the limited population size for this analysis, using a single model we computed the gait embeddings for the subjects who were both in the validation and training set.

We computed the geometric median embedding over all the control participants. For each trial for our participants with stroke, we then computed the cosine similarity between their gait and the median control embedding. This analysis used embeddings after being passed through the contrastive projection into the 16-dimensional space. We took the median similarity scores for all trials on a given day as the score for their gait at that point in rehabilitation.

\section{Results}\label{Results}

\subsection{Hyperparameter testing}\label{Hyperparameter testing}

Our HP search was a semi-structured exploration. We found the contrastive loss worked robustly on our MMC data for a wide range of network architectures. The contrastive loss on the validation data was responsive to several training parameters consistent with studies for other modalities. Specifically, a large hidden dimension, a greater number of layers, and a longer period of pretraining all tended to improve the contrastive loss. Rotary encoding also seemed to outperform learned positional encoding. While this seemed to be the trend, in a statistical analysis using multivariate analysis on the contrastive loss, two features were statistically significantly associated with an improved contrastive loss: including an additional [CLS] token and also including a next timestep prediction loss. Concerning hidden dimension size, while the contrastive loss tended to be lower with a larger hidden dimension, the next timestep prediction validation error was statistically significantly lower for smaller dimensional models. Using FCM and adding noise augmentation did not have a strong impact, and in general, seemed to worsen performance. A complete table of all the HP runs, the losses, and the downstream task performance is available in the supplementary materials (Figure~\ref{hp_table}).

\subsection{Downstream task performance}\label{Downstream task performance}

The scores on the SSL losses did not seem to predict performance on diagnosis or for laterality classification performance. The only HP that was statistically significantly linked to better classification performance was not including an addition [CLS] token, which was also statistically significantly associated with greater contrastive losses. We found across a wide range of settings that the learned representations performed well on the downstream tasks (Table~\ref{downstream_performance}). Thus, we report the aggregate statistics for classifying stroke versus control, lower-limb prosthesis user (LLPU) versus control, and laterality over all our 50 HP experiments. In general, laterality was the worst-performing task and stroke diagnosis was the best-performing task. We did not find a single best architecture that we would recommend, with many values performing well, and the raw scores are available in the supplementary materials (Figure~\ref{hp_table}).

%  |      |   Stroke |   LLPU |   Laterality |
% |:-----|---------:|-------:|-------------:|
% | mean |    0.879 |  0.774 |        0.657 |
% | std  |    0.034 |  0.043 |        0.037 |
% | max  |    0.941 |  0.842 |        0.728 |

\begin{table}
\centering
\caption{Performance statistics on downstream tasks}
\label{downstream_performance}
\begin{tabular}{p{\dimexpr 0.250\linewidth-2\tabcolsep}p{\dimexpr 0.250\linewidth-2\tabcolsep}p{\dimexpr 0.250\linewidth-2\tabcolsep}p{\dimexpr 0.250\linewidth-2\tabcolsep}}
\toprule
Metric & Stroke & LLPU & Laterality \\
\hline
Mean & 0.879 & 0.774 & 0.657 \\
Std & 0.034 & 0.043 & 0.037 \\
Max & 0.941 & 0.842 & 0.728 \\
\bottomrule
\end{tabular}
\end{table}

\subsection{Example embeddings}\label{Example embeddings}

%  One day should try this notebook based embedding https://mystmd.org/guide/cross-references#targeting-cells

\begin{figure}[!htbp]
\centering
\includegraphics[width=1\linewidth]{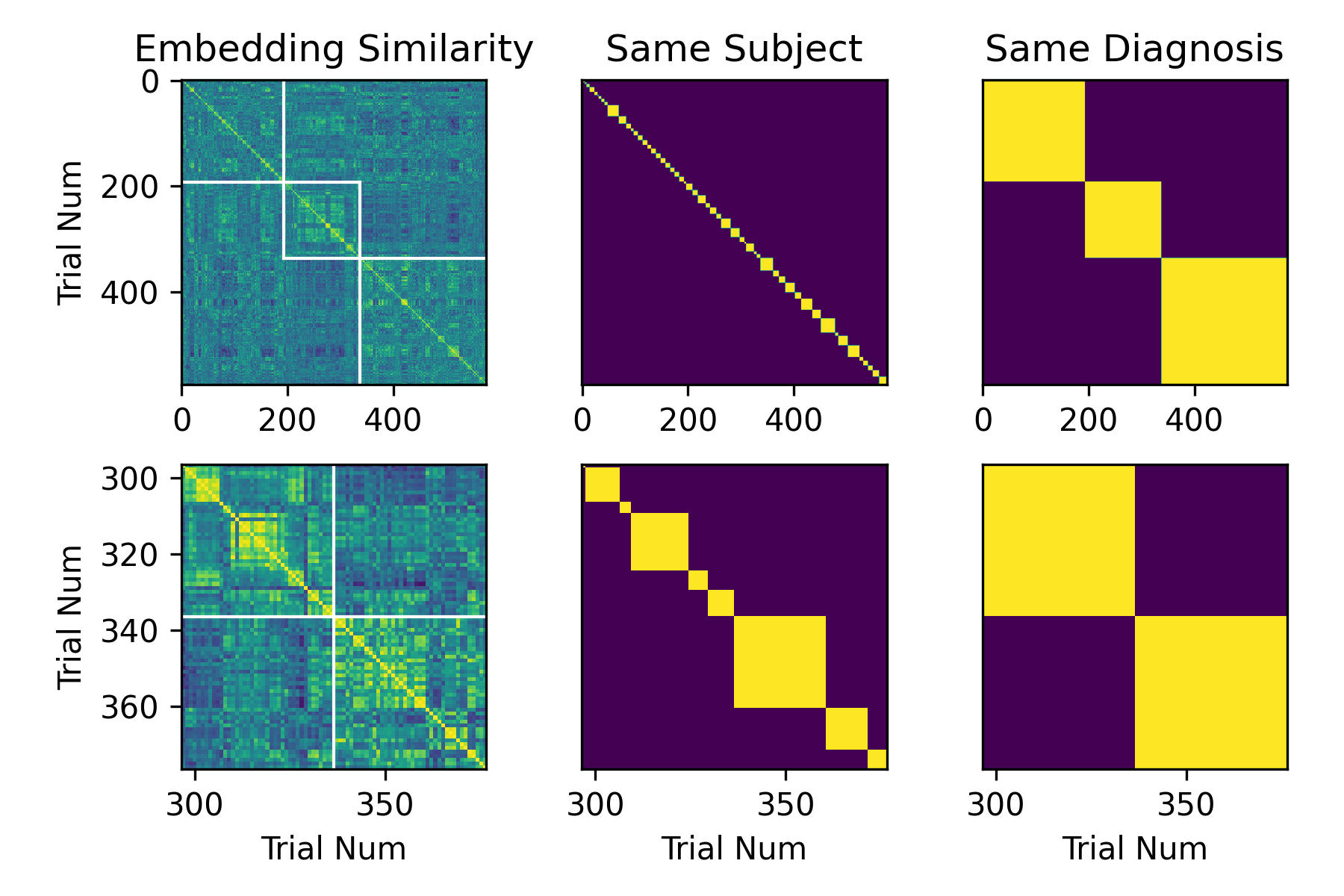}
\caption{Example similarity measured between embeddings from different trials, where each row and trial within an image is a trial. The left column shows the embedding similarity between pairs of trials. The next column shows yellow on the pairs of trials that are from the same individual. The blocks on this diagonal structure are also visible in the similarity between the embeddings from trials, showing the SSL objective learns an embedding where different trials from the same individual are similar. The last column shows whether the diagnosis was the same, with these block positions overlaid on the left column with white lines. Controls are in the bottom right corner. The second row is zoomed into the diagonal at the transition from stroke to control participants.}
\label{similiarity}
\end{figure}

To explore the learned representations, we computed the cosine similarity between the learned embeddings from different trials, ordered by participant and diagnosis (Figure~\ref{similiarity}). This visualization only includes the subset of trials from healthy controls and participants with a stroke or prothesis, our three largest groups. Despite never being supervised with subject identity, trials from the same individual were more similar than trials from different individuals. The control participants are also in the lower right and show a region that has higher overall similarity than the other regions.

\subsection{Longitudinal embeddings}\label{Longitudinal embeddings}

\begin{figure}[!htbp]
\centering
\includegraphics[width=0.625\linewidth]{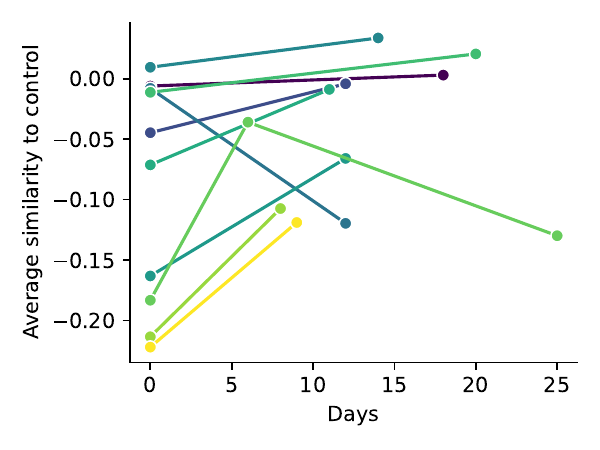}
\caption{The similarity between the average embedding of participants with a recent stroke undergoing inpatient rehabilitation and the average control participant embedding. In most cases, the average embedding for the stroke participants at each assessment becomes more aligned with the average control embedding (not visualized) throughout their rehabilitation.}
\label{longitudinal}
\end{figure}

We measured how the embedding of individuals with a recent stroke undergoing inpatient rehabilitation changed over time by measuring how aligned it was to the average control participant (Figure~\ref{longitudinal}). In most cases, the gait representation became more aligned with the average control embedding. Participants who started further from the control embedding tended to have a greater shift towards the control embedding.

There are two outliers where their embedding did not monotonically shift towards the control embedding over time. There was one participant who started fairly closely aligned to the healthy control embedding and then appeared to move away over time. In this case, they had a particularly unusual and impaired gait during the initial session with arms held fairly wide, and their gait was substantially improved at the follow-up assessment in contradiction to measuring their distances to control participants. In the second case, there was an overall trend to improvement, but an additional intermediate assessment where their gait was more aligned to control representation than at baseline or discharge. In this case, it was less apparent to which biomechanical features the gait representation was responding.

%  # Related Works

\section{Discussion}\label{Discussion}

To the best of our knowledge, the efficacy of contrastive learning of gait representations from MMC data has not been explored, although prior work showing SSL works for gait analyzed using marker-based motion capture and using HPE from monocular video suggested it was likely to be successful. Our main finding is that self-supervised learning applied to MMC-based gait analysis, using contrastive learning and next timestep prediction as pretext tasks, can learn gait representations that show promise as both diagnostic and response biomarkers.

Averaged over all our models, we obtained a cross-validated accuracy of 88\% for classifying stroke versus able-bodied control, 77\% for classifying lower-limb prosthesis users versus able-bodied control, and 66\% for identifying the laterality of their impairment. The best scores were 94\%, 84\%, and 73\% respectively. All these findings hinged on only 3 seconds of gait data, highlighting how much meaningful information SSL allows us to extract about an individual's health status from only a glimpse of their gait. The fact stroke is more easily decoded than a prosthesis user is not surprising, as in general they have a more pronounced and visible gait deviation. It is perhaps surprising that laterality is not robustly decoded and we hypothesize it is not linearly disentangled in the embedding space.

A goal of this study was to optimize the hyperparameters of SSL training to produce the best gait biomarker representation; however, this was only partially successful, because SSL loss did not robustly predict downstream task performance. The learned representations of all our models did well on the downstream task of decoding the diagnosis. The difference between models was often not much larger than the standard deviation over cross-validation folds. There was a trend towards models with smaller numbers of dimensions and not including an addition [CLS] token to have better diagnosis and laterality decoding. We anticipate the accelerating use of MMC during clinical care will produce larger datasets with richer clinical information for testing gait representations and will clarify which representations contain the most clinically meaningful information.

%  In some experiments, we also included an additional output head with a linear layer trained to minimize the cross-entropy loss over diagnosis, with the gradient from the embedding stopped before entering this head. The goal was to monitor the performance of the downstream task during training without influencing the representation. However, we found regularizing that challenge and found it did not reflect the performance of our actual trained classifiers.

We saw a more consistent influence of hyperparameters on the validation contrastive loss. The most important decision was to include an additional [CLS] token at the end of the sequence, which was fed into the contrastive loss projection layer. Including a secondary SSL loss to predict the joint angles at the next time step also improved performance. This is rather interesting as it is an entirely different loss, and this finding likely speaks to how important the dynamics of walking are to learning and understanding it. Averaged over all our experiments, pretraining on our gait laboratory dataset produced slightly lower contrastive losses, and fairly unsurprisingly, this became more pronounced for larger models. Rotary encoding with fixed frequencies generally outperformed learned positional encoding. Motivated by the success of masked modeling, we used FCM \citep{liu_towards_2023} with 20\% masking in the majority of our experiments but ultimately found it seemed to worsen performance. This might improve with higher masking levels in the future. Augmenting trajectories with noise did not seem to help, possibly because accurate biomechanical trajectories should be fairly smooth, and so noise augmentation produces a domain shift between the training and validation data. Other features we did not systematically vary may influence performance, such as the learning rate, batch size, contrastive loss temperature, and the relative weighting of the prediction loss and contrastive loss. We found the prediction loss was substantially lower for smaller hidden dimensions, although the contrastive loss tended to be lower for large hidden dimensions. We defer investigating the tradeoff between these pretext tasks on downstream tasks to future work contingent on finding a more robust performance metric for the downstream task.

Different losses might improve the learned gait representation. We generated positive samples from temporal crops of a given walking trial, which could include overlaps. Alternatively, we could define positive samples from one session from an individual or specific walking conditions using a supervised contrastive learning \citep{khosla_supervised_2020}. This could make the contrastive loss align more with the downstream task and improve on the weak association we saw between the contrastive loss and downstream task performance, but would restrict the approach to carefully annotated datasets. It is also possible to combine supervised and SSL for gait representation, such as GaitForeMer \citep{endo_gaitforemer_2022}. During pretraining on videos from healthy controls, they combined a future prediction SSL loss with an activity recognition head and then fine-tuned the activity recognition head on a clinical impairment score as well as the prediction of future activity. There are also opportunities for multimodal SSL, and our dataset also includes electromyography and inertial measurements. Recent works have shown that large pretrained language models can also be combined with motion understanding in novel ways \citep{zhang_motiongpt_2023}. Other future directions include integrating physics and biomechanical simulations into the SSL objectives, and developing algorithms that can process variable durations extending longer than 3 seconds, which is necessary to characterize gait variability. \citet{endo_gaitforemer_2022} demonstrated the benefit of clinically interpreting gait patterns from pretraining on a large monocular video dataset of mixed activities from able-bodied individuals. In addition to motion from video, there are also large datasets of movement such as AMASS \citep{mahmood_amass_2019}. We hope future work will extend SSL to learn from disparate datasets including control and clinical populations with movement estimated with different modalities to  ultimately result in a Foundation Model for gait \citep{bommasani_opportunities_2022}.

From a clinical standpoint, determining the diagnosis or laterality of impairment of an individual seen in a rehabilitation setting is of minimal practical benefit, because this is already known. Linking observational gait analysis to diagnosis and functional outcomes is routinely done by clinicians, and our experience with this motivated this study. Replicating this clinical skill with MMC gait analysis demonstrates that SSL produces gait representation that captures clinically meaningful information about the individual's health status and enables a novel diagnostic biomarker. Finding that the distance in the learned embedding space from individuals with a stroke to the average control embedding shifts towards the control embedding during inpatient rehabilitation demonstrates its potential utility as a response biomarker.

Here, we focused on both diagnostic and response biomarkers. Tantalizingly, the shape of the response biomarker resembles the proportional recovery rule in the upper extremity, where more severely impaired individuals make a large proportional recovery \citep{winters_generalizability_2015}. We also note that the idea there is a single representation of a 'healthy control gait' is a dramatic oversimplification, that serves as a simple proxy for the question of how far and individual is from their premorbid gait pattern. Additionally, more work is required to determine what biomechanical features of gait the representation is sensitive to. Ultimately, we would like to disentangle this representation into interpretable features including traditional gait measures (velocity, cadence, spatiotemporal gait parameters), diagnoses, clinical assessments, and additional holistic features capturing overall walking quality that are not captured by these prior other measures. A clear next step is to extend this work to probe for all of these features using the clinical data associated with our dataset. Mapping the learned gait representation to more interpretable features will be important for this to be useful in clinical practice and to move towards more trustworthy, explainable artificial intelligence \citep{markus_role_2021}. It will also be important to thoroughly evaluate the psychometric properties of any such response biomarkers, including reliability, validity, and sensitivity, with careful testing for any biases.

%  There was also some noise in our results from the mixture of GPUs. When running the same job on the same GPU class, the model produced identical loss functions. There would be some small variation in the cross-validated scores on the downstream tasks that we attributed to non-determinism fitting the logistic regression model. However, we would see the differences in the validation contrastive loss when running the same job on different GPU types, despite all other settings being identical. However, these differences were small relative to the difference in contrastive loss seen over the range of modeling and training configurations we used, so we do not think this would have changed our conclusions.

% %In many experiments, we had individuals walking in different conditions such as with an AFO or at a different speed than their self-selected speed, which could also define the positive samples similar to some approaches in vision to supervised contrastive learning {cite:p}`khosla_supervised_2020`.%%

One advantage of the gait signatures approach \citep{winner_discovering_2022} over our approach, is that they used autoregressive generation from their long-short term memory (LSTM) to make changes in gait signatures biomechanically interpretable. Our transformer architecture can also produce kinematics autoregressively, but must be conditioned on an entire kinematic sequence and cannot be conditioned on the contrastively learned overall gait representation. In the future, we could train a generative component conditioned on our gait representation to produce a representative gait. It will also be important to directly compare the LSTM trained with a future prediction as the SSL loss from \citet{winner_discovering_2022} to the transformer architecture used here to compare their performance decoding gait-based biomarkers. An LSTM has the advantage of autoregressive generation from an initial state, but transformers typically achieve state-of-the-art performance when training on large datasets. In preliminary studies, we also found that a temporal convolutional neural network worked well with contrastive learning, although we did not evaluate this extensively.

Finally, we hope to extend the learned gait representation to other types of biomarkers including prognostic and predictive biomarkers. The distinction between these is that a prognostic biomarker anticipates future outcomes, whereas a predictive biomarker informs the counterfactual question about outcomes based on different treatments. These are both critical for the development of precision rehabilitation \citep{cotton_letter_2022, french_precision_2022}. To develop prognostic gait-based biomarkers, we could train additional layers on these learned representations to predict future clinical outcomes. For example, predicting the discharge walking speed from the gait analysis during early rehabilitation. The development of predictive biomarkers will require additional training on data from randomized controlled trials or combining gait analysis and longitudinal, observational data with causal inference methods \citep{pearl_book_2018}, an exciting future direction.

\section{Conclusion}\label{Conclusion}

We showed that self-supervised learning from gait data with both contrastive learning and next timestep prediction produces a representation of gait that perform well on clinically relevant downstream tasks. We also found this approach works for gait analysis performed with markerless motion capture. MMC makes gait analysis much more accessible, and we envision this soon being integrated into routine clinical care, including outpatient medical visits, physical therapy visits, and at multiple points during inpatient rehabilitation. Finally, we showed that a linear layer fine-tuned on these gait representations, using only a fairly small dataset, already enables both diagnostic and response gait-based biomarkers.

\section{Ethical Considerations}\label{Ethical Considerations}

The motivation for this work is to extend gait-based biomarkers to allow more precise and frequent clinical analysis, to ultimately improve outcomes through precision rehabilitation approaches. However, developing algorithms that can infer health information from movement also carries potential risks. For example, the algorithms might be used to infer health information without the consent of the individual, or the algorithms might be used to discriminate against individuals. While the US has strict protections around sharing of health information by healthcare organizations through HIPAA, it is not clear there are yet sufficient regulations on how such ambiently sensed information could be used. An additional possible risk is that the algorithm might be used to identify individuals from their walking patterns to enable surveillance and monitoring without the consent of the individuals. As such, a concern we have around the sharing of clinical gait datasets, even if only kinematics, is that individualized gait patterns could be deanonymized and risk patient privacy. In addition, like most other artificial intelligence tools, the algorithm is not guaranteed to be 100\% correct and in fact, our current work shows this is not the case. We also have not yet investigated whether any biases influence its accuracy. We believe the benefits of this work outweigh the risks and will ultimately lead to tools that will advance rehabilitation, but strongly believe they should not be used to infer anything about an individual without their consent.

%%%%%%%%%%%%% Acknowledgements %%%%%%%%%%%%%
\section*{Acknowledgements}
\footnotesize
This work was generously supported by the Research Accelerator Program of the Shirley Ryan AbilityLab and with funding from the Restore Center P2C (NIH P2CHD101913). We thank the participants and staff of the Shirley Ryan AbilityLab for their contributions to this work.

%  Funding sources were not listed during the anonymous review.

We would like to thank the creators of Jax, Equinox, Haliax, Levanter, Optax and WandB that made it a pleasure to implement and train these models.
Large language models were used in this work. Copilot was used during code development. Some portions of the text were reviewed by ChatGPT. Grammarly was also used.
\normalsize

%%%%%%%%%%%%%%   Bibliography   %%%%%%%%%%%%%%
\bibliography{main.bib}

\clearpage
\section{Appendix}\label{Appendix}

\subsection{Participants}\label{Participants}

This study was approved by the Northwestern University Institutional Review Board.

%  This study was approved by *removed for anonymous review*.

We recruited 75 participants from the inpatient and outpatient services at our hospital as well as individuals with no gait impairments.
Of these, 18 were controls, 22 had a history of stroke, and 26 were lower limb prosthesis users, with the remainder having a variety of conditions. Ages ranged from 22 to 78. Some participants used assistive devices including orthotics, rolling walkers, or a cane, and a few participants required contact guard assistance from someone nearby for safety. The dataset included 712 trials of walking. Participants would walk multiple trials, sometimes on multiple days, and sometimes in different conditions such as with or without an ankle foot orthosis, and at different speeds. These conditions were not uniform as this was a convenience sample of participants enrolled in several projects.

For testing the representation, we classified the diagnoses as stroke, prosthesis user, able-bodied control, or other. The other category was not scored for downstream tasks. For individuals with unilateral gait impairments, we classified them as left or right.

\subsection{Markerless motion capture}\label{Markerless motion capture}

MMC data was acquired and processed using previously described system and algorithms \citep{cotton_improved_2023, cotton_markerless_2023}. For completeness, we briefly summarize this here. Multicamera data was collected with a custom system in a $7.4m \times 8m$ room with subjects walking the length of the diagonal ($11m$), producing about $7m$ of walking data. We used 10 FLIR BlackFly S GigE cameras (and 8 in several early experiments), which were synchronized using the IEEE1588 protocol and acquired data at 30 fps, with a typical spread between timestamps of less than 100µs. We used a mixture of lenses including F1.4/6mm, F1.8/12m, F1.6/4.4-11mm with lenses and positions selected to ensure at least three cameras covered the participants along the walkway.

The acquisition software was implemented in Python using the PySpin interface and includes a web-based React UI for organizing and running recording sessions. Before each experiment, calibration videos were acquired with a checkerboard ($7 \times 5$ grid of 110mm squares) spanning the acquisition volume. Extrinsic and intrinsic calibration was performed using the anipose library \citep{karashchuk_anipose_2020}. The intrinsic calibration included only the first distortion parameter.

Videos were processed using PosePipe \citep{cotton_posepipe_2022}. EasyMocap \citep{easymocap} was used for the initial multiview associations, and we implemented a 3D scene visualizer that allows clicking the subject of interest to annotate them for more precise analysis. Keypoints in the image plane were detected using a recently developed approach that is trained on numerous 3D datasets and outputs all these formats (MeTRAbs-ACAE) \citep{sarandi_learning_2022}. We used this to produce OpenPose compatible keypoints for the initial reconstruction with Easymocap and also for the biomechanical analysis. For the latter, we used the output keypoints from the MOVI dataset \citep{ghorbani_movi_2021}, which has 87 keypoint locations commonly used in optical motion capture. Importantly, this has numerous keypoints around the pelvis and torso that are not available in the more common datasets. Keypoint trajectories in 3D were reconstructed with an implicit representation approach \citep{cotton_improved_2023}. Inverse kinematics fits were performed with nimblephysics \citep{nimblephysics} and specifically with their implementation of a dual optimization process that solves for all the parameters at the same time from a set of dynamic trials \citep{werling_rapid_2022}. This used the biomechanical model we had previously optimized and validated for our MMC system and keypoints \citep{cotton_markerless_2023}

Foot contact and toe-off timing and location were acquired using a GaitRite walkway spanning the room diagonal \citep{mcdonough_validity_2001, bilney_concurrent_2003}. These were acquired as part of previously published validation studies \citep{cotton_improved_2023, cotton_markerless_2023}, and in this work were used to only select the time window between the first foot contact and last toe-off to ensure only gait data was included for SSL. From the biomechanical fits, we extracted the sagittal plane joint angles from the hip, knee, and ankles and also included back and elbow flexion.

\subsection{Clinical gait analysis laboratory dataset}\label{Clinical gait analysis laboratory dataset}

For pretraining our network, we also used marker-based data collected from our clinical gait laboratory. This dataset was previously described \citep{cotton_transforming_2022}. It includes 3541 trials from 496 participants with a wide range of diagnoses.

\subsection{Hyperparameter testing}\label{Hyperparameter testing}

\begin{figure}[!htbp]
\centering
\includegraphics[width=1\linewidth]{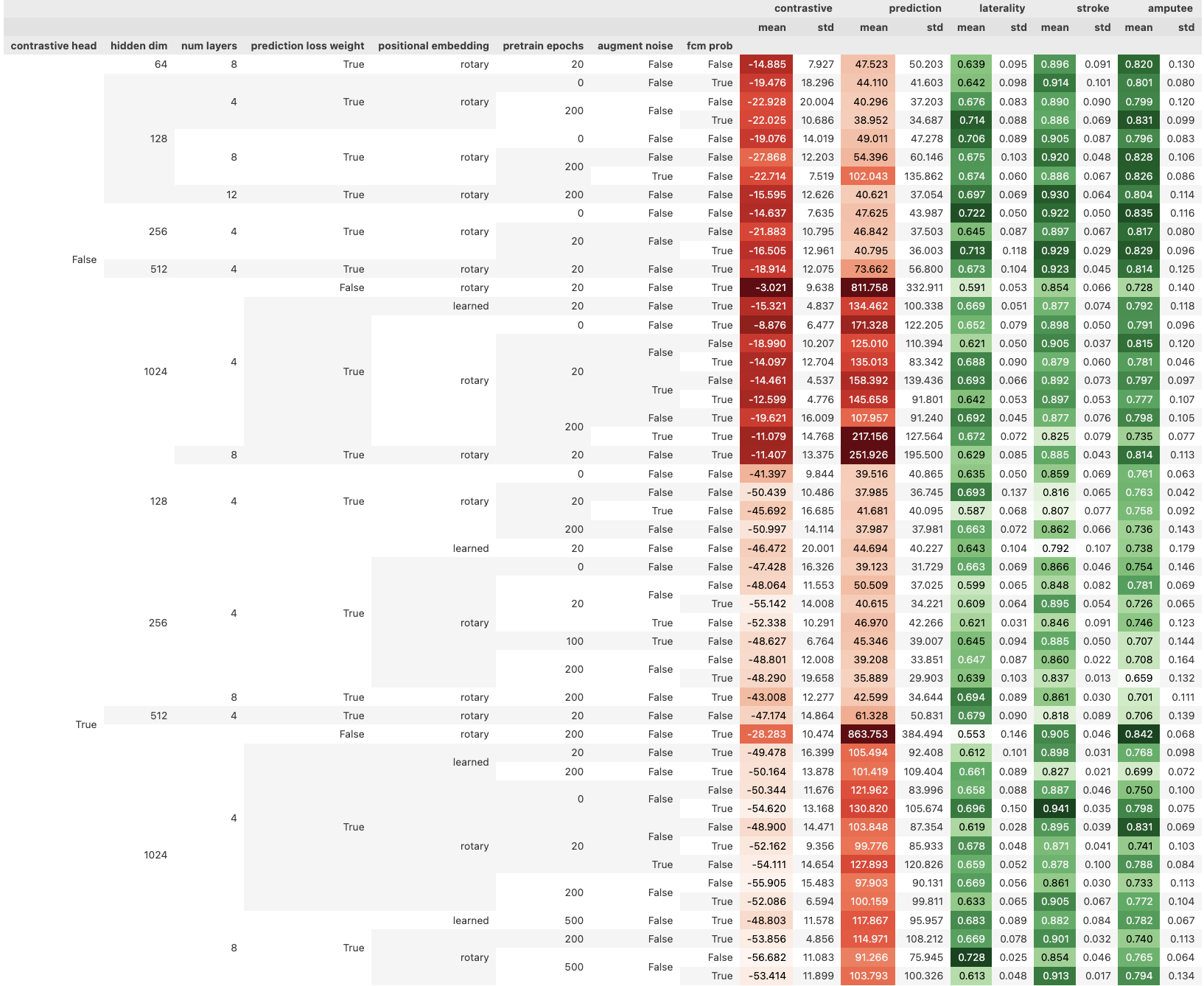}
\caption{The table shows validation loss scores and diagnosis classification sores for all the hyperparameters tested. The mean scores over cross-validation folds are colored to highlight the best scores. We also report the standard deviation over the cross validation folds in the \texttt{std} column. For the losses we used red, with darker reds being the best scores. For downstream task performance we used green, with more intense greens reflecting better classification performance.}
\label{hp_table}
\end{figure}

Figure~\ref{hp_table} shows all the scores for our hyperparameter testing.

\end{document}